\documentclass[10pt,twocolumn,letterpaper]{article}

\usepackage{cvpr}
\usepackage{times}
\usepackage{epsfig}
\usepackage{graphicx}
\usepackage{amsmath}
\usepackage{amssymb}
\usepackage{multirow, hhline}
\usepackage{color}
\usepackage{stfloats}
\usepackage{algorithm, algorithmic}
\usepackage{booktabs}
\usepackage[table,xcdraw]{xcolor}

\begin{document}
	
	\title{Spoof Trace Discovery for Deep Learning Based Explainable Face Anti-Spoofing}
	
	\author{
		{Haoyuan Zhang\textsuperscript{1,2}, Xiangyu Zhu\textsuperscript{1,2}, Li Gao\textsuperscript{3}, Jiawei Pan\textsuperscript{1,2}, Kai Pang\textsuperscript{4}, Guoying Zhao\textsuperscript{5}, Zhen Lei\textsuperscript{1,2,6,7,}\thanks{Corresponding author}}\\
		\textsuperscript{1}School of Artificial Intelligence, University of Chinese Academy of Sciences, Beijing, China\\
		\textsuperscript{2}MAIS, Institute of Automation, Chinese Academy of Sciences, Beijing, China \\
		\textsuperscript{3}China Mobile Financial Technology Co., Ltd., Beijing, China\\
		\textsuperscript{4}Guangzhou Pixel Solutions Co., Ltd., Guangzhou, China\\
		\textsuperscript{5}Center for Machine Vision and Signal Analysis, University of Oulu, Oulu, Finland\\
		\textsuperscript{6}CAIR, HKSIS, Chinese Academy of Sciences, Hong Kong, China\\
		\textsuperscript{7}SCSE, the Faculty of Innovation Engineering, M.U.S.T, Macau, China\\
		{\tt\small \{zhanghaoyuan2023, xiangyu.zhu, panjiawei2023, zhen.lei\}@ia.ac.cn} \\
		{\tt\small gaolids@chinamobile.com, pangkai@pixelall.com, guoying.zhao@oulu.fi}
	}
	
	\markboth{Journal of \LaTeX\ Class Files, Vol. 14, No. 8, August 2015}
	{Shell \MakeLowercase{\textit{et al.}}: Bare Demo of IEEEtran.cls for IEEE Journals}
	\maketitle
	
	\begin{abstract}
		With the rapid growth usage of face recognition in people's daily life, face anti-spoofing becomes increasingly important to avoid malicious attacks. Recent face anti-spoofing models can reach a high classification accuracy on multiple datasets but these models can only tell people ``this face is fake'' while lacking the explanation to answer ``why it is fake''. Such a system undermines trustworthiness and causes user confusion, as it denies their requests without providing any explanations. In this paper, we incorporate XAI into face anti-spoofing and propose a new problem termed X-FAS (eXplainable Face Anti-Spoofing) empowering face anti-spoofing models to provide an explanation. We propose SPTD (SPoof Trace Discovery), an X-FAS method which can discover spoof concepts and provide reliable explanations on the basis of discovered concepts. To evaluate the quality of X-FAS methods, we propose an X-FAS benchmark with annotated spoof traces by experts. We analyze SPTD explanations on face anti-spoofing dataset and compare SPTD quantitatively and qualitatively with previous XAI methods on proposed X-FAS benchmark. Experimental results demonstrate SPTD's ability to generate reliable explanations.
	\end{abstract}
	
	\section{Introduction}
	
	Due to the vulnerability of face recognition (FR) systems to attack, academia and industry have paid extensive attention to Face Anti-Spoofing (FAS) technology \cite{fas_survey}. Nowadays, FAS technologies \cite{flip,fmclip,fmvit,fass} have already reached a high level of defense against physical attacks such as print, replay, makeup and 3D masks, etc. However, these technologies can only answer the question ``whether the photograph provided was fake'' while lacking the evidence to support its results which brings doubts and implicit bias.

    \begin{figure}[t]
		\centering
		\includegraphics[width=\columnwidth]{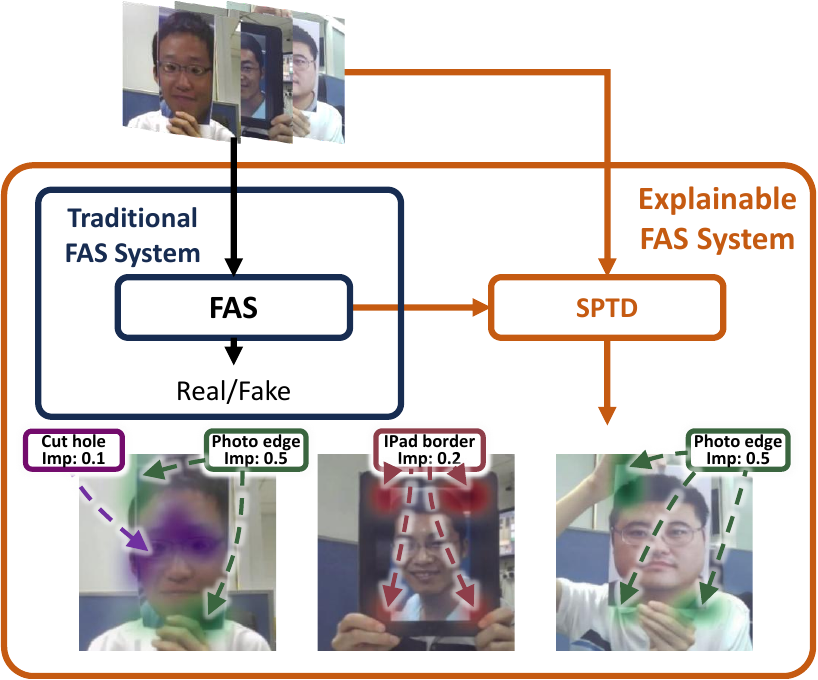}
		\caption{X-FAS method provide explanations on top of classification result. Imp indicates the \textit{importance of the concept}.}
		\label{fig:front}
		\vspace{-10pt}
	\end{figure}
    
	When a FR system rejects an image for security reasons, it is considered necessary to provide an explanation. Without such explanation, the interaction can become frustrating and uncomfortable for users, which lack transparency and trustworthiness. Thus an X-FAS (eXplainable Face Anti-Spoofing) system is advocated to provide user-friendly results by generating explanations based on FAS classification models, which is the goal of this paper. We believe X-FAS is crucial as it can make the FAS system more trustworthy and significantly enhance the user experiences.
	
	The field of eXplainable Artificial Intelligence (XAI) \cite{doshi2017towards,jacovi2021formalizing,wang2024roadmap,dwivedi2023explainable} has emerged to demystify the inner workings of black-box models and offer insights into their decision-making processes. Traditional XAI methods focused on providing a single explanation for a given input, typically in the form of heatmaps that highlight key regions. However, recent advancements in XAI \cite{poeta2023concept,sun2024explain} have underscored the importance of concept discovery which can provide not only heatmaps but also the corresponding activated concepts. These breakthroughs in XAI techniques lay a solid foundation for exploring the field of X-FAS.
	
	
	In this paper, we introduce SPTD (SPoof Trace Discovery), an X-FAS method designed to discover spoof concepts and provide explanations for attack images. Given a well-trained FAS model, SPTD can discover spoof concepts from a given FAS dataset and analyze the importance of each concept. With the help of discovered concepts, SPTD can mark the attention region of each concept if the input image is judged as a fake sample during inference. Notably, the whole process no need to updating the FAS model, thus keep the original performance. Examples are shown in Figure~\ref{fig:front}, SPTD find multiple activated concepts in attack images and provide the corresponding attention regions. To evaluate X-FAS methods, we present an X-FAS benchmark that includes 13 spoof types and 777 samples, aimed at assessing the quality of explanations generated by X-FAS techniques. Experiments demonstrate that SPTD can identify key spoof concepts and provide heatmaps corresponding to these concepts in FAS tasks, thus enhancing the trustworthiness of the system for users. Our main contributions can be described as follows:
	\begin{enumerate}
		\item We propose a new problem termed X-FAS (eXplianable Face Anti-Spoofing) that generates reliable explanations on top of FAS classification results and introduce an X-FAS method SPTD (SPoof Trace Discovery) which can discover spoof concepts and provide attention region of each concept in attack images.
		\item We propose an X-FAS benchmark with expert annotations to evaluate the quality of explanations generated by X-FAS methods.
	\end{enumerate}
	
	The quantitative results on the X-FAS benchmark, along with the qualitative analysis of the generated explanations demonstrate the efficacy of SPTD. These results highlight its ability to provide reliable explanations, thereby enhancing the trustworthiness of FAS models.
	
	\section{Related works}\label{related_works}
	
	
	\subsection{Face anti-spoofing (FAS)}
	
	FAS aims to differentiate between real and fake facial images, preventing deception in facial recognition systems. Over the years, FAS has advanced across various dimensions, including model architecture, task formulation, and training paradigms. In terms of model architecture, FAS has evolved from Convolution Neural Networks (CNNs) \cite{resnet, convnext} to Vision Transformers (ViT) \cite{vit, swin} and now to large models \cite{ifas, tffas}. Regarding task formulation, FAS has expanded from single-modal RGB recognition \cite{casiafasd, replayattack} to multimodal tasks \cite{cefa} incorporating infrared and depth images, as well as from within-domain \cite{oc_scmnet, huang2023ldcformer, huang2021face} to cross-domain tasks \cite{huang2023towards, huang2022adaptive}. In terms of training paradigms, FAS has shifted from single-image input \cite{iadg, sda, udgfas} to pre-trained vision-language contrastive learning \cite{flip, cfpl}. These advancements have made FAS more powerful, but the opacity of black-box neural networks still limits the trustworthiness of FAS systems. Although previous methods varies from training paradigms, they use same backbones (e.g. ResNet \cite{resnet}, ConvNext \cite{convnext}, ViT \cite{vit}). Thus, we propose a framework to provide additional explanations for CNN-based FAS methods to enhance system trustworthiness.
	
	
	\subsection{Explainable AI (XAI)}
	
	Explainable AI (XAI) aims to improve the interpretability and transparency of deep learning models by providing human-understandable explanations for their decisions. In computer vision, gradient-based methods such as Grad-CAM \cite{gradcam}, Ablation-CAM \cite{ablationcam}, and related approaches \cite{gradcampp, lrp} leverage activation and gradient information within the model to attribute decisions to specific regions of the input image (e.g., highlighting an airplane when classifying an image as a plane). In contrast, perturbation-based methods like RISE \cite{rise} treat the network as a black box, relying solely on input-output pairs. These methods systematically perturb the input image and observe the resulting changes in the model’s output to infer decision attribution. Additionally, concept-based approaches such as ACE \cite{ace} and CRAFT \cite{craft} automatically decompose a single decision into multiple interpretable concepts, enhancing human understanding. In the context of FAS, identifying multiple spoof traces from a single prediction aligns well with the objectives of concept-based methods, making them particularly suitable for improving model explainability in FAS.
	
	\begin{figure}[t]
		\centering
		\includegraphics[width=\columnwidth]{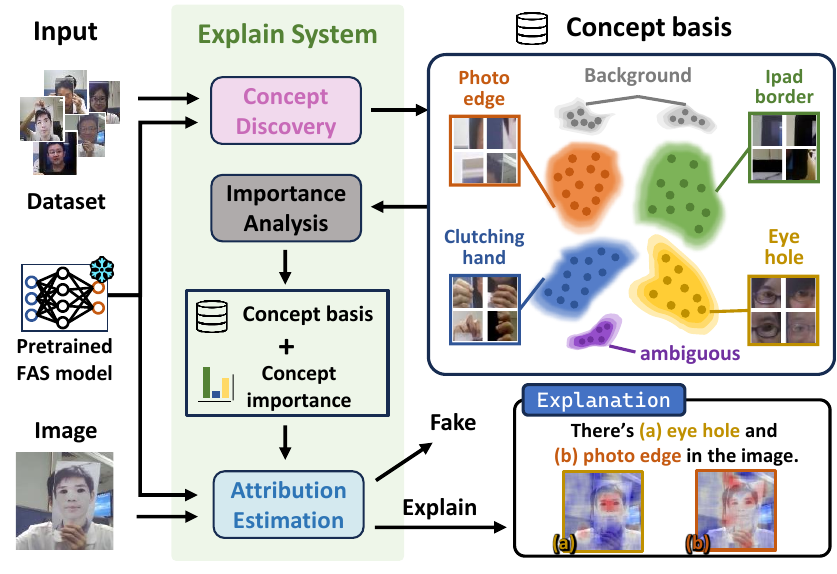}
		\caption{The overall pipeline of the proposed SPTD framework is depicted in this figure. SPTD include three parts: concept discovery, importance analysis and attribution estimation.}
		\label{fig:method_structure}
		\vspace{-10pt}
	\end{figure}
	
	\subsection{XAI + FAS}
	
	Early studies on XAI for FAS aimed to identify key differences between attack and genuine images \cite{neto2022causality}. Jourabloo \textit{et al.} \cite{jourabloo2018face} and Liu \textit{et al.} \cite{liu2020disentangling} employed disentanglement techniques to separate spoof noise from facial features, facilitating both detection and image reconstruction. Wang \textit{et al.} \cite{wang2022disentangled} and Fang \textit{et al.} \cite{fang2022learnable} utilized frequency decomposition to improve generalization across datasets by capturing distinct attack patterns. Pan \textit{et al.} \cite{pan2022attention} incorporated Grad-CAM visualization and textual explanations to enhance interpretability through attention-based training. Although these methods offer objective explanations that are faithful to the model, the need to train from scratch limits their practical applicability. Recently, Zhang \textit{et al.} proposed I-FAS \cite{ifas}, which incorporates a Large Language Model into the FAS process to generate textual explanations. However, since the generated explanation is still part of the black-box model’s output, it carries the same unreliability and does not enhance the trustworthiness of the FAS system. Based on the above, the method proposed in this paper ensures the faithfulness of the generated explanations through XAI methods, without the need for additional training. This enables SPTD to effectively enhance the practical trustworthiness of the FAS system for users.
	
	
	\section{Method}\label{method}
	
	In Face Anti-Spoofing (FAS) tasks, physical attack samples typically contain multiple spoof traces. Therefore, we propose SPTD (SPoof Trace Discovery) to provide a user-friendly explanation, which can necessarily extract various spoof traces from a single attack sample. Given a well trained FAS model, we first discover spoof trace concepts from a group of selected attack data. Secondly, we analyze importance of each concepts through perturbations. Finally, given a single spoof sample, SPTD show activated spoof trace concepts and mark their regions respectively. Thus, we separate SPTD into three parts which are concept discovery, importance analysis and attribution estimation. The whole pipeline can be seen in Figure~\ref{fig:method_structure}.
	
	
	
	\subsection{Preliminaries}
	
	Consider a general supervised FAS task setting, where the original dataset $(\boldsymbol{x}_1, \cdots, \boldsymbol{x}_N) \in \mathcal{X}^N \in \mathbb{R}^{N \times D}$ contains $N$ input images and $(y_1, \cdots, y_n) \in \mathcal{Y}^N$ their associated labels. We are given a well trained predictor $\boldsymbol{f}: \mathcal{X} \rightarrow \mathcal{Y}$ which maps the input $\boldsymbol{x}$ to the predicted class $y = \boldsymbol{f} ( \boldsymbol{x} ) $. We decompose the neural network $\boldsymbol{f}$ into two components $\boldsymbol{g}$ and $\boldsymbol{h}$ where $\boldsymbol{g}$ maps input $\boldsymbol{x}$ to intermediate logits $\boldsymbol{g} ( \boldsymbol{x} )$ and the second maps the intermediate logtis $\boldsymbol{g} ( \boldsymbol{x} )$ to output $\boldsymbol{h} ( \boldsymbol{g} ( \boldsymbol{x} ) )$. The original function $\boldsymbol{f}$ can be reconstructed as $\boldsymbol{f} = \boldsymbol{h} \circ \boldsymbol{g}$. The decomposition of $\boldsymbol{f}$ can occur at any layer of the network, though it is often chosen to be the last layer before classifier, as it contains more semantic information.
	
	\begin{figure*}[t]
		\centering
		\includegraphics[width=\textwidth]{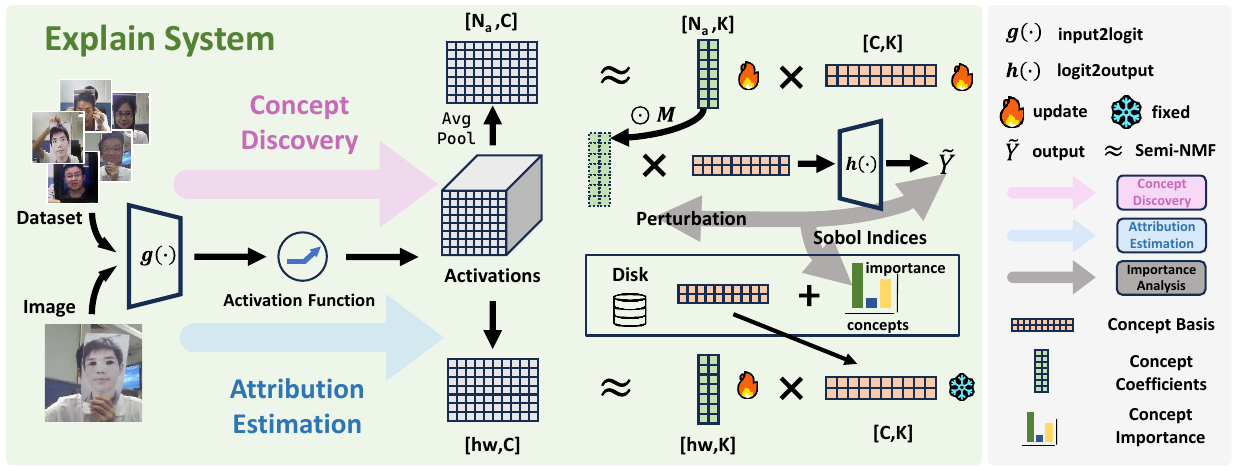}
		\caption{The detailed explain system of SPTD is illustrated above. We first identify concepts from a given dataset to construct a concept basis. Next, we assess the importance of each concept by perturbing its corresponding coefficients. Finally, leveraging the discovered concept basis and importance scores, SPTD estimates the attribution for a given image.}
		\label{fig:method_detail}
		\vspace{-10pt}
	\end{figure*}
	
	\subsection{Concept discovery}
	
	The process of concept discovery is illustrated in Figure~\ref{fig:method_detail} with \textcolor[RGB]{255,105,180}{\textbf{pink arrow}}. Firstly, we select a subset of images $\mathcal{X}_{sub} \in \mathbb{R}^{N' \times D}$ from the original dataset $\mathcal{X}^N$ for concept discovery, ensuring that the selected images share common characteristics, such as all being attack samples or belonging to a specific spoof type (e.g., print attacks). In this paper, we randomly sample $r$ frames from each attack video in $\mathcal{X}^{N}$ to form $\mathcal{X}_{sub}$ where $r$ is a hyper parameter. We assume $\pi(\cdot)$ is a filter function to create candidate spoof traces. It can be a straightforward crop and resize function to create sub-regions candidates. In the SPTD method, $\pi(\cdot)$ generates candidate spoof traces by uniformly sampling patches in both vertical and horizontal directions. Feed $\mathcal{X}_{sub}$ into $\pi(\cdot)$ to obtain an auxiliary dataset $\mathbf{X} \in \mathbb{R}^{N_a \times D}$ which contains all candidate spoof traces.
	
	To automatically discover spoof trace concepts from auxiliary dataset $\mathbf{X}$, we feed it to the network to obtain activation $\mathbf{A} = \boldsymbol{g} (\mathbf{X}) \in \mathbb{R}^{N_a \times hw \times C}$ where $hw$ indicates the shape of activation map and $C$ indicates the number of channel. We apply Semi-NMF (Semi Non-negative Matrix Factorization) \cite{seminmf} to factorize activation maps since the non-negative constraint on the coefficients brings better interpretability while remain the capability to process negative values. Semi-NMF decompose the average pooled activations $\mathbf{\bar{A}} = AvgPool(\mathbf{A}) \in \mathbb{R}^{N_a \times C}$ into a product of concept coefficients $\mathbf{U} \in \mathbb{R}^{N_a \times K}$ and concept basis $\mathbf{W} \in \mathbb{R}^{C \times K}$ by solving:
	\begin{align}
		(\mathbf{U}, \mathbf{W}) = \mathop{\arg\min}\limits_{\mathbf{U} \geq 0, \mathbf{W}} \Vert \mathbf{\bar{A}} - \mathbf{U} \mathbf{W}^\intercal \Vert^2_F, \label{eq:seminmf}
	\end{align}
	where K indicates the number of concepts one wish to discover and $ \Vert \cdot \Vert^2_F $ denotes the Frobenius norm.
	
	Following Ding \textit{et al.} \cite{seminmf}, we solve the above objective by iteratively updating $\mathbf{U}$ and $\mathbf{W}$. Specifically, $\mathbf{W}$ is the discovered spoof trace concepts where each column $\mathbf{W}_k \in \mathbb{R}^{C}$ corresponds to a single spoof trace concept. These concepts will be utilized in the subsequent importance analysis and attribution estimation process.
	
	\subsection{Importance analysis}
	
	The process of importance analysis is illustrated in Figure~\ref{fig:method_detail} with \textcolor[RGB]{105,105,105}{\textbf{gray arrow}}. We adopt sobol indices \cite{sobol} to estimate the importance of each spoof trace concept. Given the selected $N'$ images $\mathcal{X}_{sub}$ and the discovered concepts basis $\mathbf{W}$. We first feed images from $\mathcal{X}_{sub}$ to the network to obtain activations $\mathbf{A} = \boldsymbol{g} (\mathcal{X}_{sub}) \in \mathbb{R}^{N' \times hw \times C}$. Then we decompose the activation in each position into several concept coefficients $\mathbf{U}$ corresponding to the basis $\mathbf{W}$ by solving similar objective in Equation \ref{eq:seminmf}:
	\begin{align}
		(\mathbf{U}^{(i, j)}, \mathbf{W}) = \mathop{\arg\min}\limits_{\mathbf{U}^{(i, j)} \geq 0, \mathbf{W}} \Vert \mathbf{A}^{(i, j)} - \mathbf{U}^{(i, j)} \mathbf{W}^\intercal \Vert^2_F, \label{eq:seminmf2}
	\end{align}
	where $\mathbf{W}$ is fixed based on the values obtained during the concept discovery process, and $\mathbf{A}^{(i, j)} \in \mathbb{R}^{N' \times C}$ represents the feature vector at position $(i, j)$ within the spatial dimensions $(h, w)$. The result $\mathbf{U}^{(i, j)}  \in \mathbb{R}^{N' \times K}$ indicates the concept coefficients in position $(i, j)$.
	
	Formally, a common way to estimate the importance of a concept $k$ is to measure the variations of the model's output $\boldsymbol{h}(\mathbf{U}\mathbf{W}^\intercal)$ when concept coefficient $\mathbf{U}_{(1, k)}, \cdots, \mathbf{U}_{(N', k)}$ undergo meaningful perturbations. We generate random perturbation mask $\mathbf{M} \sim \mathcal{U}[ 0, 1 ]^{K}$ and reconstruct the perturbed activation $\mathbf{\widetilde{A}} = (\mathbf{U} \odot \mathbf{M}) \mathbf{W}^\intercal$. Thus, the perturbed output can be denoted as $\mathbf{\widetilde{Y}} = \boldsymbol{h}(\mathbf{\widetilde{A}})$. Simply understanding, the model output will vary substantially when perturbing an important concept, while a less relevant concept will have little to no impact. The importance of concept $k$ can be written as:
	\begin{align}
		\mathcal{S}_k = \frac{\mathbb{E}_{\mathbf{M}_{\sim k}} (\mathbb{V}_{\mathbf{M}_k} ( \mathbf{\widetilde{Y}} \vert \mathbf{M}_{\sim k} ) )}{\mathbb{V} ( \mathbf{\widetilde{Y}} ) },
	\end{align}
	where $\mathbb{E}$ indicates expectation and $\mathbb{V}$ indicates variance. We use Sobol Sequence as the random generator of perturbation mask $\mathbf{M}$.

	\subsection{Attribution estimation}\label{attribution}
	
	The process of attribution estimation is illustrated in Figure~\ref{fig:method_detail} with \textcolor[RGB]{0,191,255}{\textbf{blue arrow}}. After concept discovery, we get $K$ concept basis $\mathbf{W} \in \mathbb{R}^{C \times K}$ from the the selected subset of $N'$ images, which serves as an approximation of the spoof trace concepts present in the original dataset containing $N$ images. Given an image $\boldsymbol{x}$ (from the original dataset or any other data source), we factorize the activations $\mathbf{A} = \boldsymbol{g} (\boldsymbol{x}) \in \mathbb{R}^{hw \times C}$ with the fixed concept basis $\mathbf{W}$ through Equation \ref{eq:seminmf2} and get the corresponding concept coefficients $\mathbf{U} \in \mathbb{R}^{hw \times K}$. Specifically, we can regard $\mathbf{U}^{(x, y)}_{k}$ as the importance of concept $k$ at $(x, y)$ position of the activation map and the $\mathbf{U}_{k}$ can be seen as an activation map of concept $k$. In this way, we factorize a single input image into several concepts and its activation map. With the help of concept coefficients $\mathbf{U}$, we can use previous attribution method to enhance the estimation of concept attribution. We introduce C-RISE (Concept RISE) which modified RISE (Randomized Input Sampling for Explanation) \cite{rise} to work for concept attribution.
	
%
	
	\begin{figure*}[t]
		\centering
		\includegraphics[width=\linewidth]{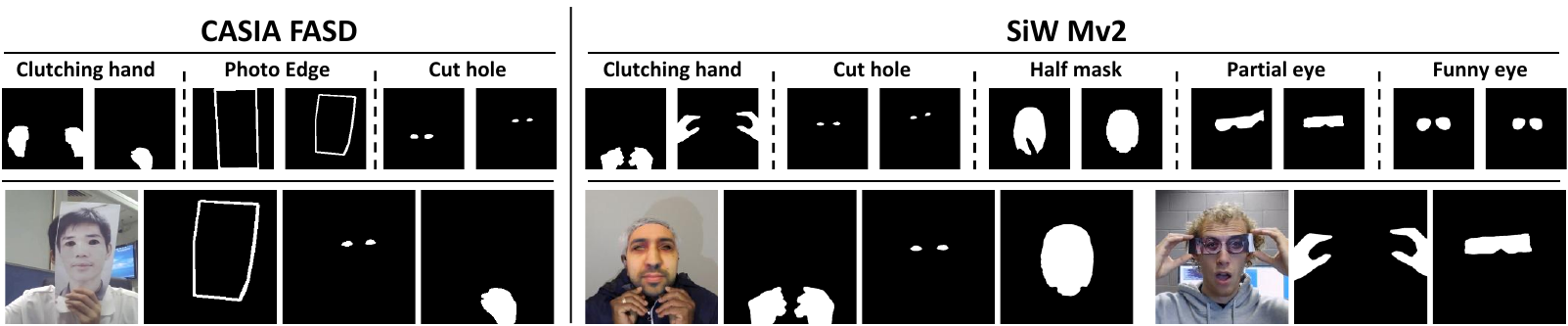}
		\caption{Sample visualization of X-FAS benchmark. We show samples of annotation mask and attack images with multiple spoof traces.}
		\label{fig:benchmark}
		\vspace{-10pt}
	\end{figure*}
	
	
	RISE \cite{rise} treats the target model as a black box, requiring only input-output pairs without needing access to the internal inference process. Given an image $\mathbf{x}$, RISE applies random masks $\mathbf{M}$ to each pixel and computes the expected output influence under perturbations:  
	\begin{align}  
		S(\mathbf{x}) = \mathbb{E}_{\mathbf{M}} \left[ \mathbf{M} \cdot \mathbf{f}(\mathbf{x} \odot \mathbf{M}) \right]  
	\end{align}  
	where $\odot$ denotes element-wise multiplication.
	
	To estimate the pixel importance of a specific concept $k$, we focus on the coefficients $\mathbf{U}_{k}$ instead of the classification logits. The estimation in C-RISE can be formulated as:  
	\begin{align}  
		S_k(\mathbf{x}) = \mathbb{E}_{\mathbf{M}} \left[ \mathbf{M} \cdot \text{Avg}( \mathbf{Fac}_k ( \mathbf{g}(\mathbf{x} \odot \mathbf{M}) ) ) \right]  
	\end{align} 
	where $\mathbf{Fac}_k$ denotes Semi-NMF factorization result of concept $k$ and $\mathbf{Avg}$ represents average pooling along the spatial dimension.
	
	\begin{table}[t]
		\centering
		\caption{Detailed information of the proposed X-FAS benchmark.}
		\label{tab:benchmark}
		\resizebox{\columnwidth}{!}{%
			\begin{tabular}{c|c|cccc}
				\hline
				Dataset & Spoof Type                                                   & Subjects            & Number of frame      & Spoof Traces   & Number of mask \\ \hline
				\multirow{6}{*}{CASIA-FASD} &
				\multirow{3}{*}{Print} &
				\multirow{3}{*}{16} &
				\multirow{3}{*}{16} &
				Photo edge &
				16 \\
				&                                                              &                     &                      & Cut hole       & 8              \\
				&                                                              &                     &                      & Clutching hand & 16             \\ \cline{2-6} 
				& \multirow{2}{*}{Replay}                                      & \multirow{2}{*}{8}  & \multirow{2}{*}{8}   & Ipad border    & 8              \\
				&                                                              &                     &                      & Clutching hand & 8              \\ \cline{2-6} 
				& Total                                                        & 24                  & 24                   & All            & 56             \\ \hline
				\multirow{23}{*}{SiW-Mv2} &
				\multirow{2}{*}{\begin{tabular}[c]{@{}c@{}}Makeup\\ Impersonation\end{tabular}} &
				\multirow{2}{*}{1} &
				\multirow{2}{*}{2} &
				Eye brows &
				2 \\
				&                                                              &                     &                      & Makeup eye     & 2              \\ \cline{2-6} 
				& \begin{tabular}[c]{@{}c@{}}Makeup\\ Obfuscation\end{tabular} & 9                   & 18                   & Makeup mark    & 18             \\ \cline{2-6} 
				& Mannequin                                                    & 40                  & 80                   & Model\_head    & 80             \\ \cline{2-6} 
				& \multirow{3}{*}{Half Mask}                                   & \multirow{3}{*}{66} & \multirow{3}{*}{66}  & Half mask      & 66             \\
				&                                                              &                     &                      & Cut hole       & 63             \\
				&                                                              &                     &                      & Clutching hand & 31             \\ \cline{2-6} 
				& \multirow{2}{*}{Paper Mask}                                  & \multirow{2}{*}{17} & \multirow{2}{*}{33}  & Paper mask     & 34             \\
				&                                                              &                     &                      & Clutching hand & 34             \\ \cline{2-6} 
				&
				\multirow{3}{*}{\begin{tabular}[c]{@{}c@{}}Transparent\\ Mask\end{tabular}} &
				\multirow{3}{*}{58} &
				\multirow{3}{*}{58} &
				Transparent Mask &
				58 \\
				&                                                              &                     &                      & Cut hole       & 58             \\
				&                                                              &                     &                      & Clutching hand & 3              \\ \cline{2-6} 
				& \multirow{2}{*}{Partial Eye}                                 & \multirow{2}{*}{56} & \multirow{2}{*}{104} & Partial eye    & 104            \\
				&                                                              &                     &                      & Clutching hand & 104            \\ \cline{2-6} 
				&
				\multirow{2}{*}{\begin{tabular}[c]{@{}c@{}}Funny\\ Eye Glasses\end{tabular}} &
				\multirow{2}{*}{176} &
				\multirow{2}{*}{176} &
				Funny eye &
				176 \\
				&                                                              &                     &                      & Funny Eyeball  & 14             \\ \cline{2-6} 
				& \multirow{2}{*}{Partial Mouth}                               & \multirow{2}{*}{29} & \multirow{2}{*}{58}  & Paper mouth    & 58             \\
				&                                                              &                     &                      & Clutching hand & 58             \\ \cline{2-6} 
				& \multirow{2}{*}{Paper Glasses}                               & \multirow{2}{*}{75} & \multirow{2}{*}{132} & Paper glass    & 132            \\
				&                                                              &                     &                      & Clutching hand & 3              \\ \cline{2-6} 
				& \multirow{2}{*}{Silicone}                                    & \multirow{2}{*}{14} & \multirow{2}{*}{26}  & Silicon mask   & 26             \\
				&                                                              &                     &                      & Cut hole       & 26             \\ \cline{2-6} 
				& Total                                                        & 541                 & 753                  & All            & 1150           \\ \hline
			\end{tabular}%
		}
	\vspace{-10pt}
	\end{table}
	
	\section{Benchmark}\label{benchmark}
	
	In order to evaluate the explanation quality generated by X-FAS methods, we use expert annotated spoof traces that can repeatably evaluate multiple explainable methods in an unbiased way. Following the annotation manner of Kondapaneni et al. \cite{kondapaneni2024less} which proposed an expert-defined birds feature dataset to evaluate explainable methods for birds classification, we introduce an X-FAS benchmark for testing X-FAS methods which can measure the accuracy of generated explanation at a fine-grained level.
	
	\subsection{Fine-grained explanation dataset}
	
	The benchmark data is spoof images from the CASIA-FASD \cite{casiafasd} and SiW-Mv2 \cite{siwmv2} dataset. The detailed information of the proposed X-FAS benchmark is illustrated in Table~\ref{tab:benchmark}. CASIA-FASD consists of two attack types: print attacks and replay attacks, while SiW-Mv2 includes a diverse set of attack types, such as various mask attacks and model head attacks. For both CASIA-FASD and SiW-Mv2, we automatically select frames from original videos using a pretrained CLIP \cite{clip} model, following the process outlined in Algorithm~\ref{alg:filter}. Our approach aims to maximize the dissimilarity between sampled frames from each video while ensuring that a face is detected in every selected frame. By analyzing obvious shared spoof traces with human experts' knowledge, we annotate the regions of each spoof traces, finally forming 1206 spoof trace masks of 777 attack images in total to produce targets of X-FAS methods. Samples in the X-FAS benchmark are visualized in Figure~\ref{fig:benchmark}.
	
	\begin{algorithm}[h]
		\small
		\renewcommand{\algorithmicensure}{\textbf{Input:}}
		\renewcommand{\algorithmicrequire}{\textbf{Output:}}
		\caption{Filter frames from video}
		\label{alg:filter}
		\begin{algorithmic}[1]
			\ENSURE	Video $\mathcal{V}$ = \{$F_1, F_2, \cdots, F_N$, where $F_i$ indicates the $i$ th frame\}, Frame number $l$,
			Random sample iteration $Iter$, Pretrained CLIP model $\mathcal{M}_{clip}$, Retina face detection model $\mathcal{M}_{face}$
			\REQUIRE Selected frames $\mathcal{S}$ 
			\STATE	$\mathcal{V}'$ = \{$F_1, F_2, \cdots, F_{N'}$, where $M_{face}(F_i)$ detects a face\}
			\STATE	Extract CLIP Feature $\mathcal{E}$ of all frames from $\mathcal{V}'$ : \\
			$\mathcal{E} = \mathcal{M}_{clip}(\mathcal{V}') \in \mathbb{R}^{N \times D}$
			\STATE	$max\_sim \gets -\infty$
			\STATE	$L \gets None$
			\FORALL{$t = 1, \cdots, Iter$}
			\STATE	$sim \gets 0$
			\STATE	Random choose $l$ frames from $\mathcal{V}'$ :\\
			$L^{sample} = random\_choice(F_1, F_2, \cdots, F_{N'})$
			\FORALL{$i = 1, \cdots, l - 1$}
			\FORALL{$j = i + 1, \cdots, l$}
			\STATE 	$sim += 1 - \frac{\mathcal{E}_{L^{sample}_i} \cdot \mathcal{E}_{L^{sample}_j}}{||\mathcal{E}_{L^{sample}_i}|| \cdot ||\mathcal{E}_{L^{sample}_j}||}$
			\ENDFOR
			\ENDFOR
			\IF{$max\_sim < sim$}
			\STATE	$max\_sim = sim$
			\STATE	$L = L_{sample}$
			\ENDIF
			\ENDFOR
			\STATE	$\mathcal{S} \gets $  \{$F_{L_1}, F_{L_2}, \cdots, F_{L_l}$\}
			\RETURN $\mathcal{S}$
		\end{algorithmic}
	\end{algorithm}
	
	\subsection{Evaluation protocol}\label{protocol}
	
	In the X-FAS benchmark, CASIA-FASD and SiW-Mv2 subsets should be evaluated separately. To ensure reliable testing, the model used should perform well on the target dataset (e.g., when evaluating the SiW-Mv2 subset, the pretrained model should be trained using the SiW-Mv2 intra-protocol). With a well-trained model, explanations can be generated using X-FAS methods, and their quality can be assessed by computing metrics that compare the generated explanations to the ground truth annotations for each image in the benchmark. If multiple annotation masks exist for an image, the final metric should be computed as the average across all masks.
	
	\subsection{Evaluation metric}\label{niou}
	
	\textit{Intersection over Union} (IoU) is a widely used metric and can also be applied in the X-FAS benchmark. However, directly averaging the IoU of multiple spoof traces is unfair, as the theoretical maximum IoU varies due to differences in the pixel count of ground truth annotation masks. To address this, we propose a fairer metric derived from IoU, termed \textit{normalized Intersection over Union} (nIoU). Given a annotated mask $\mathbf{M}_G$ and an explanation $\mathbf{M}_I$, we first obtain a processed explanation mask $\mathbf{M}^x_I$ by setting the top $x$ percent of values to 1 and the rest to 0. The nIoU metric is then formulated as follows:
	\begin{equation}
		\begin{aligned}
			nIoU(\mathbf{M}_G, \mathbf{M}_I, x) = \frac{IoU(\mathbf{M}_G, \mathbf{M}^x_I) * max(x, y)}{min(x, y)} \in [0, 1]
		\end{aligned}	
	\end{equation}
	where $y$ is the useful pixel percentage of annotated traces $\mathbf{M}_G$. $\frac{min(x, y)}{max(x, y)}$ is the optimum value of $IoU(\mathbf{M}_G, \mathbf{M}^x_I)$
	
	\begin{table*}[t]
		\centering
		\caption{Quantitative result on X-FAS benchmark. When calculating IoU and nIoU, we consider the top 30\% pixels as the explanation mask. The evaluation protocol follows \cite{kondapaneni2024less}.}
		\label{tab:expertres}
		\resizebox{\textwidth}{!}{%
			\begin{tabular}{c|cc|ccccccc}
				\hline
				Dataset                     & \multicolumn{2}{c|}{Method}                                                                                 & GradCAM         & GradCAM++       & EigenGradCAM    & AblationCAM     & RandomCAM & RISE            & \cellcolor[HTML]{C0C0C0}SPTD (Ours)    \\ \hline
				\multirow{6}{*}{CASIA-FASD} & \multicolumn{1}{c|}{\multirow{2}{*}{Print}}                                                          & IoU  & 0.0648          & 0.0622          & 0.0681          & 0.0968          & 0.0585    & 0.0801          & \cellcolor[HTML]{C0C0C0}\textbf{0.1078} \\
				& \multicolumn{1}{c|}{}                                                                                & nIoU & 0.3354          & 0.3811          & 0.4268          & 0.4688          & 0.2585    & 0.3707          & \cellcolor[HTML]{C0C0C0}\textbf{0.6141} \\ \hhline{~|---------} 
				& \multicolumn{1}{c|}{\multirow{2}{*}{Replay}}                                                         & IoU  & 0.1530          & 0.1023          & 0.0606          & 0.1598          & 0.0962    & 0.1840          & \cellcolor[HTML]{C0C0C0}\textbf{0.2244} \\
				& \multicolumn{1}{c|}{}                                                                                & nIoU & 0.2545          & 0.1358          & 0.1169          & 0.2882          & 0.1959    & 0.4048          & \cellcolor[HTML]{C0C0C0}\textbf{0.4837} \\ \hhline{~|---------} 
				& \multicolumn{1}{c|}{\multirow{2}{*}{Average}}                                                        & IoU  & 0.0942          & 0.0756          & 0.0656          & 0.1178          & 0.0760    & 0.1110          & \cellcolor[HTML]{C0C0C0}\textbf{0.1467} \\
				& \multicolumn{1}{c|}{}                                                                                & nIoU & 0.3084          & 0.2993          & 0.3235          & 0.4086          & 0.3026    & 0.3679          & \cellcolor[HTML]{C0C0C0}\textbf{0.5706} \\ \hline
				\multirow{24}{*}{SiW-Mv2}   & \multicolumn{1}{c|}{\multirow{2}{*}{\begin{tabular}[c]{@{}c@{}}Makeup\\ Impersonation\end{tabular}}} & IoU  & \textbf{0.0295} & \textbf{0.0295} & \textbf{0.0295} & \textbf{0.0295} & 0.0171    & \textbf{0.0295} & \cellcolor[HTML]{C0C0C0}\textbf{0.0295} \\
				& \multicolumn{1}{c|}{}                                                                                & nIoU & \textbf{1.0000} & \textbf{1.0000} & \textbf{1.0000} & \textbf{1.0000} & 0.5103    & \textbf{1.0000} & \cellcolor[HTML]{C0C0C0}\textbf{1.0000} \\ \hhline{~|---------} 
				& \multicolumn{1}{c|}{\multirow{2}{*}{\begin{tabular}[c]{@{}c@{}}Makeup\\ Obfuscation\end{tabular}}}   & IoU  & 0.1211          & \textbf{0.1319} & \textbf{0.1319} & 0.1243          & 0.0568    & 0.1019          & \cellcolor[HTML]{C0C0C0}0.1318          \\
				& \multicolumn{1}{c|}{}                                                                                & nIoU & 0.9156          & 0.9999          & \textbf{1.0000} & 0.9372          & 0.4123    & 0.7423          & \cellcolor[HTML]{C0C0C0}0.9991          \\ \hhline{~|---------} 
				& \multicolumn{1}{c|}{\multirow{2}{*}{Mannequin}}                                                      & IoU  & 0.4317          & 0.4580          & 0.5019          & 0.4595          & 0.2828    & 0.3923          & \cellcolor[HTML]{C0C0C0}\textbf{0.5173} \\
				& \multicolumn{1}{c|}{}                                                                                & nIoU & 0.7028          & 0.7478          & 0.8105          & 0.7481          & 0.4686    & 0.6405          & \cellcolor[HTML]{C0C0C0}\textbf{0.8453} \\ \hhline{~|---------} 
				& \multicolumn{1}{c|}{\multirow{2}{*}{Half Mask}}                                                      & IoU  & 0.2189          & 0.2418          & 0.2494          & 0.2318          & 0.0815    & 0.1648          & \cellcolor[HTML]{C0C0C0}\textbf{0.2629} \\
				& \multicolumn{1}{c|}{}                                                                                & nIoU & 0.8243          & 0.8678          & 0.8803          & 0.8583          & 0.3185    & 0.7429          & \cellcolor[HTML]{C0C0C0}\textbf{0.9619} \\ \hhline{~|---------} 
				& \multicolumn{1}{c|}{\multirow{2}{*}{Paper Mask}}                                                     & IoU  & 0.2373          & 0.2362          & 0.2422          & 0.2322          & 0.1022    & 0.1497          & \cellcolor[HTML]{C0C0C0}\textbf{0.2802} \\
				& \multicolumn{1}{c|}{}                                                                                & nIoU & 0.6297          & 0.6110          & 0.6663          & 0.6153          & 0.3142    & 0.5631          & \cellcolor[HTML]{C0C0C0}\textbf{0.9296} \\ \hhline{~|---------} 
				& \multicolumn{1}{c|}{\multirow{2}{*}{\begin{tabular}[c]{@{}c@{}}Transparent\\ Mask\end{tabular}}}     & IoU  & 0.1805          & 0.2750          & 0.2783          & 0.2251          & 0.1255    & 0.1423          & \cellcolor[HTML]{C0C0C0}\textbf{0.2795} \\
				& \multicolumn{1}{c|}{}                                                                                & nIoU & 0.7060          & 0.9830          & 0.9887          & 0.8409          & 0.4927    & 0.5741          & \cellcolor[HTML]{C0C0C0}\textbf{0.9922} \\ \hhline{~|---------} 
				& \multicolumn{1}{c|}{\multirow{2}{*}{Partial Eye}}                                                    & IoU  & 0.1358          & 0.1271          & 0.1344          & 0.1517          & 0.0853    & 0.1917          & \cellcolor[HTML]{C0C0C0}\textbf{0.2122} \\
				& \multicolumn{1}{c|}{}                                                                                & nIoU & 0.6298          & 0.6153          & 0.6317          & 0.6691          & 0.3199    & 0.7533          & \cellcolor[HTML]{C0C0C0}\textbf{0.8147} \\ \hhline{~|---------} 
				& \multicolumn{1}{c|}{\multirow{2}{*}{\begin{tabular}[c]{@{}c@{}}Funny\\ Eye Glasses\end{tabular}}}    & IoU  & 0.1077          & 0.1121          & 0.1120          & 0.1080          & 0.0601    & 0.1059          & \cellcolor[HTML]{C0C0C0}\textbf{0.1123} \\
				& \multicolumn{1}{c|}{}                                                                                & nIoU & 0.9448          & 0.9977          & 0.9950          & 0.9513          & 0.5296    & 0.9330          & \cellcolor[HTML]{C0C0C0}\textbf{0.9985} \\ \hhline{~|---------} 
				& \multicolumn{1}{c|}{\multirow{2}{*}{\begin{tabular}[c]{@{}c@{}}Partial\\ Mouth\end{tabular}}}        & IoU  & 0.1885          & 0.1690          & 0.1679          & 0.1951          & 0.0962    & 0.2007          & \cellcolor[HTML]{C0C0C0}\textbf{0.2438} \\
				& \multicolumn{1}{c|}{}                                                                                & nIoU & 0.6971          & 0.6283          & 0.6225          & 0.7194          & 0.3312    & 0.7218          & \cellcolor[HTML]{C0C0C0}\textbf{0.8852} \\ \hhline{~|---------} 
				& \multicolumn{1}{c|}{\multirow{2}{*}{\begin{tabular}[c]{@{}c@{}}Paper\\ Glasses\end{tabular}}}        & IoU  & 0.0914          & 0.0927          & 0.0927          & 0.0927          & 0.0406    & 0.0914          & \cellcolor[HTML]{C0C0C0}\textbf{0.0944} \\
				& \multicolumn{1}{c|}{}                                                                                & nIoU & 0.9742          & 0.9924          & 0.9924          & 0.9811          & 0.4223    & 0.9598          & \cellcolor[HTML]{C0C0C0}\textbf{0.9964} \\ \hhline{~|---------} 
				& \multicolumn{1}{c|}{\multirow{2}{*}{Silicone}}                                                       & IoU  & 0.2082          & 0.2393          & \textbf{0.2738} & 0.2219          & 0.1192    & 0.1822          & \cellcolor[HTML]{C0C0C0}0.2711          \\
				& \multicolumn{1}{c|}{}                                                                                & nIoU & 0.7236          & 0.7708          & 0.8009          & 0.7332          & 0.3631    & 0.6508          & \cellcolor[HTML]{C0C0C0}\textbf{0.8119} \\ \hhline{~|---------} 
				& \multicolumn{1}{c|}{\multirow{2}{*}{Average}}                                                        & IoU  & 0.1739          & 0.1859          & 0.1939          & 0.1848          & 0.0974    & 0.1645          & \cellcolor[HTML]{C0C0C0}\textbf{0.2154} \\
				& \multicolumn{1}{c|}{}                                                                                & nIoU & 0.8103          & 0.8516          & 0.8644          & 0.8388          & 0.4306    & 0.7868          & \cellcolor[HTML]{C0C0C0}\textbf{0.9345} \\ \hline
			\end{tabular}%
		}
	\vspace{-10pt}
	\end{table*}
	
	\section{Experiments}
	
	
	To demonstrate the effectiveness of SPTD, we primarily evaluate it on the Face Anti-Spoofing (FAS) task using the proposed X-FAS benchmark. We compare SPTD with several representative XAI methods, analyze the discovered spoof concepts, and visualize fine-grained explanations to highlight its strengths in interpreting FAS models. In addition, to further verify the fidelity of SPTD, we conduct supplementary experiments on a general vision task (ImageNet classification) by comparing it with CRAFT \cite{craft}. This evaluation provides an objective perspective on how well SPTD reflects the model’s true decision-making process.
	
	
	\subsection{Qualitative Results on X-FAS Benchmark}\label{sec:xfas}
	
	\subsubsection{Baselines}
	
	We consider three categories of XAI methods: gradient-based methods, perturbation-based methods, and concept-based methods. Gradient-based methods include GradCAM \cite{gradcam}, GradCAM++ \cite{gradcampp}, EigenGradCAM \cite{eigencam}, AblationCAM \cite{ablationcam}, RandomCAM (code from \cite{pytorchcam}). For the perturbation-based category, we adopt RISE \cite{rise} as a representative method. The concept-based method is SPTD proposed in this paper. Following the evaluation protocol in \cite{kondapaneni2024less}, we calculate mean IoU and mean nIoU (detail in Section \ref{niou}) metric on different spoof types of X-FAS benchmark.
	
	\subsubsection{Experimental settings}
	
	We adopt the widely recognized FLIP \cite{flip} method and train two FLIP-V models separately on the CASIA-FASD \cite{casiafasd} and SiW-Mv2 \cite{siwmv2} datasets. The model trained on CASIA-FASD achieves an HTER of 0.11\% on the test split, while the model trained on SiW-Mv2 achieves 2.49\%, demonstrating their strong performance. All previous XAI methods, along with the proposed SPTD method, are evaluated using these two well-trained FLIP-V models. For all XAI methods, we select the last ResBlock as the target layer. In SPTD, we set the number of concepts to $K = 15$ and use vanilla estimation. The subset for concept discovery and importance analysis is generated by randomly selecting two frames from each video in the original dataset. For a fair comparison, we use vanilla attribution, which has been employed by previous XAI methods. Vanilla attribution follows Collins \textit{et al.} \cite{dff} by marking the attention regions of each concept through scaling the activation maps (keeping 10\% of the maximum value and setting the rest to zero) and mapping them back to the input shape, as activation maps preserve spatial correlations with the input image.
	
	\subsubsection{Results}
	
	Table \ref{tab:expertres} presents the overall results on the X-FAS benchmark. The results indicate that among all previous XAI methods, AblationCAM and RISE outperform other methods on the CASIA-FASD dataset, while EigenGradCAM achieves the best results on the SiW-Mv2 dataset. EigenGradCAM excels in three test items related to Makeup and Silicone, whereas our proposed SPTD method outperforms on the remaining nineteen scenarios. Furthermore, in terms of average IoU and nIoU, SPTD achieves the highest results among all methods. These findings demonstrate that SPTD surpasses previous XAI methods on X-FAS benchmark, highlighting the superior quality of its explanations.
	
	
	\subsection{Visualization and Analysis}
	
	With the same experimental settings in Section~\ref{sec:xfas}, we visualize the discovered concepts and the generated explanations of SPTD on CASIA-FASD \cite{casiafasd} dataset.
	
	\subsubsection{Spoof Concept and Explanation Visualization}
	
\begin{figure}[t]
		\centering
		\includegraphics[width=0.9\linewidth]{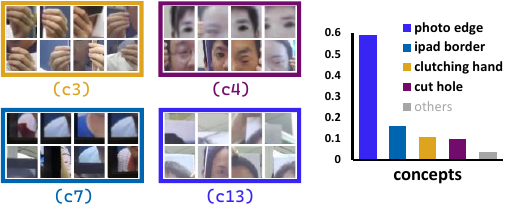}
		\caption{Discovered spoof concepts in CASIA-FASD dataset. Concept c3, c4, c7 and c13 are top four important concepts.}
		\label{fig:casia_concepts}
		\vspace{-10pt}
	\end{figure}

	We visualize the discovered top four important spoof concepts on CASIA-FASD dataset with their analyzed importance in Figure~\ref{fig:casia_concepts}. As we can see, the discovered concepts can be easily understood by users since they are represented as multiple patches. We further attempt to summarize these four discovered concepts and find clear semantic meanings where c3, c4, c7 and c13 expresses \textit{clutching hand}, \textit{cut hole}, \textit{photo edge} and \textit{iPad border} respectively.
	
\begin{figure}[ht]
		\centering
		\includegraphics[width=0.9\linewidth]{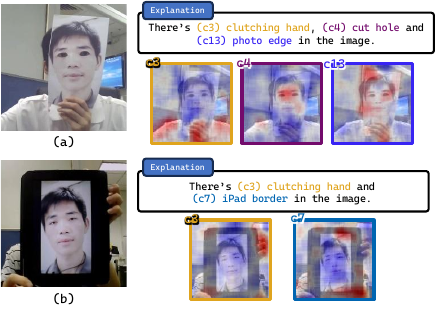}
		\caption{Explanations on CASIA-FASD samples of concepts in Figure~\ref{fig:casia_concepts}. Sample (a) activated concept c3 (\textit{clutching hand}), c4(\textit{cut hole}) and c13 (\textit{photo edge}) while (b) activated c3 (\textit{clutching hand}) and c7 (\textit{iPad border}). Heatmaps show pixel level attention region of each activated concept.}
		\label{fig:casia_explanations}
		\vspace{-15pt}
	\end{figure}

	With the help of these discovered concepts, SPTD can generate fine-grained explanations which mark attention regions of activated corresponding concepts using C-RISE attribution estimation mentioned in Section~\ref{attribution}, as shown in Figure~\ref{fig:casia_explanations}. In the explanation of Figure~\ref{fig:casia_explanations}(a), SPTD find activated concepts c3, c4, c13 and mark the specific attention region in red which is consistent with concepts in Figure~\ref{fig:casia_concepts}. While in Figure~\ref{fig:casia_explanations}(b), SPTD only find activated concept c3 and c7.

	\begin{figure*}[t]
		\centering
		\includegraphics[width=0.95\textwidth]{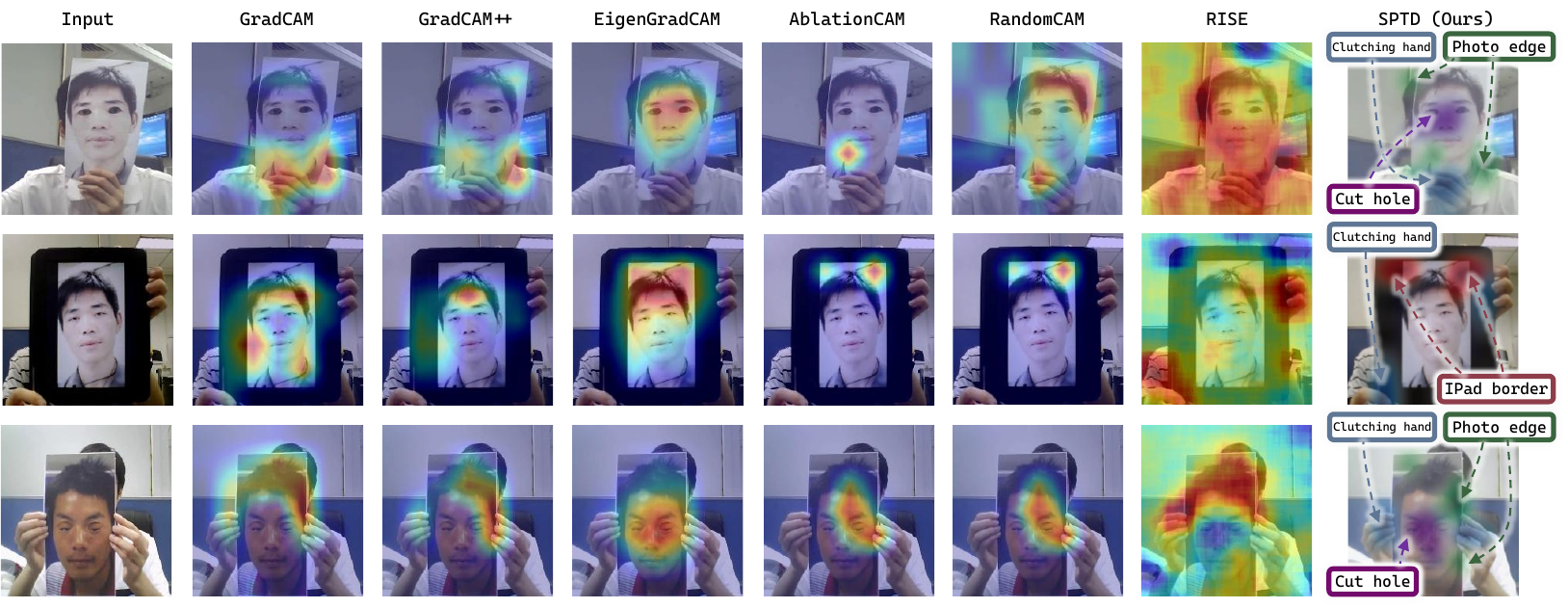}
		\caption{Visualization of multiple XAI methods on CASIA-FASD samples. We compare multiple previous XAI methods with SPTD, previous methods give a single heatmap while SPTD gives multiple attention regions of corresponding activated spoof concepts.}
		\label{fig:expert_com}
	\vspace{-10pt}
	\end{figure*}

	The result shows that SPTD has the ability to discover spoof concepts which are easy to be understood by users and provide corresponding attention regions of each concept on top of face anti-spoofing models.
	
	\subsubsection{Explanation Comparison}
	
	Figure~\ref{fig:expert_com} shows the visualization comparison between multiple XAI methods and SPTD (using vanilla estimation) on CASIA-FASD \cite{casiafasd} dataset. Previous XAI methods show a single heatmap where some cover the whole face missing finer-level information and some pay attention to partial spoof traces. In contrast, SPTD provides multiple heatmaps where each of them corresponds to a specific activated spoof concept. These results prove that with the help with SPTD explanations, we can surely increase the trustworthiness to face anti-spoofing system users.
	
	\subsection{Fidelity Evaluation}
	
	To further demonstrate the fidelity of SPTD, we adopt two metrics to compare SPTD and CRAFT \cite{craft} (both are concept-based method) on several random selected classes of ImageNet. \textit{Deletion} and \textit{Insertion} are a pair of the most popular evaluation methods to check an explanation’s fidelity. \textit{Deletion} measures the drop in model confidence by progressively removing the most important features. A stronger explanation is expected to cause a larger drop in confidence. In contrast, \textit{Insertion} evaluates the confidence increase by gradually adding the same features back where larger confidence increase is preferred. For both Deletion and Insertion metrics, we calculate the area under the curve (AUC) which represents the quality of explanation method. A lower deletion AUC and higher insertion AUC is appreciated.
	
	We compare SPTD and CRAFT on class 101 and 413 of ImageNet using the same pretrained ResNet18 model. Since CRAFT also analyze importance of each discovered concepts, the top three important concepts are fetched to calculate fidelity metrics.
	
%
	


    \begin{table}[ht]
	 	\centering
	 	\caption{Fidelity results on class 101 and 413 of ImageNet. Top@k indicates the k th important concept. \textit{Ins} and \textit{Del} indicates \textit{Insertion} and \textit{Deletion} metric correspondingly.}
	 	\label{tab:imagenet}
	 	\vspace{10pt}
	 	\begin{minipage}{0.49\linewidth}
	 		\centering
	 		\resizebox{\columnwidth}{!}{%
	 			\begin{tabular}{ccccc}
	 				\hline
	 				Class                                     & metrics                                                                                       & top@k & CRAFT   & \begin{tabular}[c]{@{}c@{}}SPTD\\ (Ours)\end{tabular} \\ \hline
	 				\multicolumn{1}{c|}{\multirow{8}{*}{101}} & \multicolumn{1}{c|}{\multirow{4}{*}{\begin{tabular}[c]{@{}c@{}}Ins\\ (↑)\end{tabular}}} & Top1  & 0.46335 & \textbf{0.48351}                                            \\
	 				\multicolumn{1}{c|}{}                     & \multicolumn{1}{c|}{}                                                                         & Top2  & 0.44794 & \textbf{0.48701}                                            \\
	 				\multicolumn{1}{c|}{}                     & \multicolumn{1}{c|}{}                                                                         & Top3  & 0.36773 & \textbf{0.41768}                                            \\
	 				\multicolumn{1}{c|}{}                     & \multicolumn{1}{c|}{}                                                                         & Avg   & 0.42634 & \textbf{0.46273}                                            \\ \cline{2-5} 
	 				\multicolumn{1}{c|}{}                     & \multicolumn{1}{c|}{\multirow{4}{*}{\begin{tabular}[c]{@{}c@{}}Del\\ (↓)\end{tabular}}}  & Top1  & 0.23237 & \textbf{0.21507}                                            \\
	 				\multicolumn{1}{c|}{}                     & \multicolumn{1}{c|}{}                                                                         & Top2  & 0.24420 & \textbf{0.21876}                                            \\
	 				\multicolumn{1}{c|}{}                     & \multicolumn{1}{c|}{}                                                                         & Top3  & 0.30649 & \textbf{0.30107}                                            \\
	 				\multicolumn{1}{c|}{}                     & \multicolumn{1}{c|}{}                                                                         & Avg   & 0.26102 & \textbf{0.24497}                                            \\ \hline
	 			\end{tabular}%
	 		}
	 	\end{minipage}
	 	\hfill
	 	\begin{minipage}{0.49\linewidth}
	 		\centering
	 		\resizebox{\columnwidth}{!}{%
	 			\begin{tabular}{ccccc}
	 				\hline
	 				Class                                     & metrics                                                                                       & top@k & CRAFT   & \begin{tabular}[c]{@{}c@{}}SPTD\\ (Ours)\end{tabular} \\ \hline
	 				\multicolumn{1}{c|}{\multirow{8}{*}{413}} & \multicolumn{1}{c|}{\multirow{4}{*}{\begin{tabular}[c]{@{}c@{}}Ins\\ (↑)\end{tabular}}} & Top1  & 0.48587 & \textbf{0.48697}                                            \\
	 				\multicolumn{1}{c|}{}                     & \multicolumn{1}{c|}{}                                                                         & Top2  & 0.38627 & \textbf{0.45782}                                            \\
	 				\multicolumn{1}{c|}{}                     & \multicolumn{1}{c|}{}                                                                         & Top3  & 0.31657 & \textbf{0.43796}                                            \\
	 				\multicolumn{1}{c|}{}                     & \multicolumn{1}{c|}{}                                                                         & Avg   & 0.39624 & \textbf{0.46092}                                            \\ \cline{2-5} 
	 				\multicolumn{1}{c|}{}                     & \multicolumn{1}{c|}{\multirow{4}{*}{\begin{tabular}[c]{@{}c@{}}Del\\ (↓)\end{tabular}}}  & Top1  & 0.14864 & \textbf{0.14722}                                            \\
	 				\multicolumn{1}{c|}{}                     & \multicolumn{1}{c|}{}                                                                         & Top2  & 0.23849 & \textbf{0.17079}                                            \\
	 				\multicolumn{1}{c|}{}                     & \multicolumn{1}{c|}{}                                                                         & Top3  & 0.30824 & \textbf{0.19949}                                            \\
	 				\multicolumn{1}{c|}{}                     & \multicolumn{1}{c|}{}                                                                         & Avg   & 0.23179 & \textbf{0.17250}                                            \\ \hline
	 			\end{tabular}%
	 		}
	 	\end{minipage}
	 \vspace{-10pt}
	 \end{table}
	
	 Table~\ref{tab:imagenet} presents the quantitative results of CRAFT and SPTD. As shown, SPTD outperforms CRAFT in terms of fidelity across all settings, indicating that SPTD is more effective in discovering concepts and estimating attributions. Notably, CRAFT is only applicable when activation maps are non-negative. In contrast, SPED is free from this constraint, and the results demonstrate its ability to generate superior explanations without relying on such restrictions.

	\section{Conclusion}
	
	In this paper, we propose a  new problem termed X-FAS to provide reliable face anti-spoofing results by generating explanations on top of face anti-spoofing classification results to cope with the vulnerability of black-box models. We introduce SPTD (SPoof Trace Discovery), an X-FAS method which can discover spoofing concepts that is easy to be understood by users and provide a heatmap of activated concepts of attack images. To evaluate the quality of X-FAS methods, we present an X-FAS benchmark with expert annotations on two FAS datasets. In our experiments, both quantitative and qualitative results show the efficacy and reliability of SPTD. We hope that this work will guide further efforts in the research for X-FAS which can eliminate user's doubts of face anti-spoofing models and make it more transparent, trustworthy and effective.
	
	\section*{Acknowledgement}
	
	This work was supported in part by Chinese National Natural Science Foundation Projects U23B2054, 62276254, 62306313, the Beijing Science and Technology Plan Project Z231100005923033, Beijing Natural Science Foundation L221013, the Science and Technology Development Fund of Macau Project 0140/2024/AGJ, and InnoHK program.
	
	{\small
		\bibliographystyle{ieee}
		\bibliography{ref}
	}

\end{document}